\documentclass{article}

\PassOptionsToPackage{sort&compress,numbers}{natbib}

     \usepackage[final]{neurips_2019}

\usepackage[utf8]{inputenc} %
\usepackage[T1]{fontenc}    %
\usepackage{hyperref}       %
\usepackage{url}            %
\usepackage{booktabs}       %
\usepackage{amsfonts}       %
\usepackage{amsmath}       %
\usepackage{nicefrac}       %
\usepackage{microtype}      %
\usepackage{booktabs}
\usepackage{cleveref}
\usepackage{tikz} %
\usepackage{pgfplots}
\usetikzlibrary{positioning,calc,arrows.meta} %
\usepackage{amsmath} %
\usepackage{float}
\usepackage{xcolor}

\title{Think Again Networks and the Delta Loss}

\author{%
  Alexandre Salle \qquad Marcelo Prates \\
  Institute of Informatics \\
  Universidade Federal do Rio Grande do Sul \\
  \texttt{alex@alexsalle.com,morprates@inf.ufrgs.br} \\
}

\begin{document}

\maketitle

\begin{abstract}
This short paper introduces an abstraction called Think Again Networks (ThinkNet) which can be applied to any state-dependent function (such as a recurrent neural network).
\end{abstract}

\section{Think Again Networks}

Given an arbitrary state-dependent function $F(x, s)$ where $x$ is the input and $s$ is the state, a ThinkNet $\textit{TN}$ is given by:

\begin{equation}
  \textit{TN}(F, x, s, t) = \begin{cases} F(x, s) &\mbox{if } t = 1 \\
  F(x, M([\textit{TN}(F, x, s, 1);\dots;\textit{TN}(F, x, s, t-1)])) &\mbox{otherwise} \end{cases}
\end{equation}

where $s$ is an initial state, $t \geq 1$ is a ThinkNet timestep, $M([z_1;...;z_{t-1}])$ is a arbitrary \emph{mixing} function (could be a static function or a trainable neural network) which combines all previous outputs into a new state. Rather than optimize a loss $\mathcal{L}(F(.,.))$ directly, one optimizes $\mathcal{L}(TN(F,.,.,T))$ instead.

\begin{figure}[H]
\centering
\begin{tikzpicture}

    \def\unitdist{2.5}

    \foreach \i in {1,...,4}
    {
        \node[rectangle, draw, minimum height=1.25cm, rounded corners=.25cm, fill=red!20!white!80, rotate=90] (unrol-\i) at (\unitdist*\i,0) {$F(x,s)$};
    }
    
    \node[rectangle, draw, minimum width=1cm, minimum height=1cm, rounded corners=.25cm, fill=red!20!white!80] (X) at (\unitdist*2.5,-2.5) {$x$};
    
    \path[->, >=stealth,  thick] (X) edge [bend left=12] node [xshift=-0cm,yshift=-0.45cm] {}  (unrol-1);
    \path[->, >=stealth,  thick] (X) edge [bend left=6] node [xshift=-0cm,yshift=-0.45cm] {}  (unrol-2);
    \path[->, >=stealth,  thick] (X) edge [bend left=-6] node [xshift=-0cm,yshift=-0.45cm] {}  (unrol-3);
    \path[->, >=stealth,  thick] (X) edge [bend left=-12] node [xshift=-0cm,yshift=-0.45cm] {}  (unrol-4);
    
    \node[rectangle, draw, minimum width=.75cm, minimum height=.75cm, rounded corners=.25cm, fill=red!20!white!80] (loss-unit-4) at (\unitdist*4,2.5) {$\mathcal{L}$};
    \node[] (loss-4) at (\unitdist*4,3.5) {$\mathcal{L}$};
    \draw [->, >=stealth,  thick] (loss-unit-4) -- (loss-4);
    \draw [ ->, >=stealth,  thick, red!50!black!50] (unrol-4) -- (loss-unit-4);
    
    \foreach \i in {1,...,3}
    {
        \node[rectangle, draw, minimum width=.75cm, minimum height=.75cm, rounded corners=.25cm, fill=red!20!white!80] (M-unit-\i) at (\unitdist*\i+0.75,2.5) {$M$};
    }
    
    \foreach \i in {1,...,3}
    {
         \foreach \j in {\i,...,3}
         {
            \path[ ->, >=stealth, thick, red!50!black!50] (unrol-\i) edge [bend left=12] node [xshift=-0cm,yshift=-0.45cm] {}  (M-unit-\j);
         }
    }
    
     \path[ ->, >=stealth, thick, blue!50!black!50] (0.5,2) edge [bend right=15] node [xshift=.75cm,yshift=-0.25cm] {$s^{(0)}$}  ($(unrol-1.center) + (-0.75,0)$);
    
    \foreach \i in {1,...,3}
    {
        \pgfmathparse{\i+1}
        \xdef\iplusone{\pgfmathresult}
        
        \path[ ->, >=stealth, thick, blue!50!black!50] (M-unit-\i) edge [bend right=15] node [xshift=.75cm,yshift=-0.25cm] {$s^{(\i)}$}  ($(unrol-\iplusone) + (-0.75,-0.5)$);
    }

\end{tikzpicture}
\caption{Diagram for a ThinkNet (implemented on top of a state-dependent function $F(x,s)$) running for 4 timesteps. At $t=0$, the network is initialized with a state $s^{(0)}$. For subsequent timesteps $t > 0$, the mixing function $M$ computes the next state $s^{(t+1)}$ given all outputs produced at the previous timesteps $t,t-1,\dots,1$. Finally, a loss $\mathcal{L}$ is computed at the end of the pipeline.}
\label{fig:thinknet}
\end{figure}
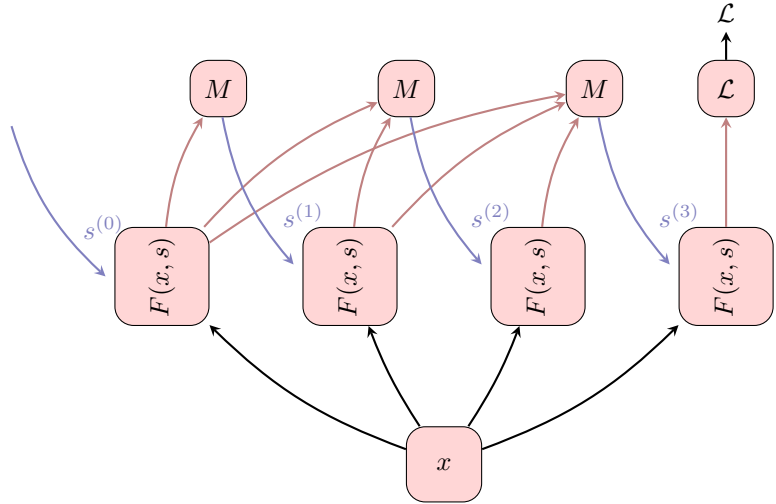

Essentially, the Think Again abstraction extends any state-dependent function $F(x,s)$ by adding an additional level of recurrence to it. The idea is to run $F(x,s)$ repeatedly $t$ times (each run corresponds to a ThinkNet timestep), and have a mixing function $M$ prepare the initial state of the next execution. In order to do so, $M$ can be defined as a function over all previous outputs produced by the network. The metaphor for the Think Again abstraction is a human problem-solver completing a task in many passes: it is often useful to think about the problem at hand an additional time after one has already processed it and updated his ``mental state'' ($s$) about it. Conceptually, any problem can be approached in this framework: humans can learn to reason about all their previous answers ($z_1,z_2,\dots,z_{t-1}$) when attacking a problem an additional time (to do that, we train our own ``mixing function'' $M$). This makes sense both from the perspective of human problem-solvers and deep learning models, which may benefit from \emph{knowing in advance an internal representation of the problem while the are processing it}. In the context of sentence/sequence encoding, one may benefit from knowing in advance how a sentence/sequence ends while reading it.

Although related works in multi-step processing (similar to our ThinkNet timesteps) \citep{niehues2016pre,chatterjee2016fbk,zhou2017neural,junczys2017exploration,xia2017deliberation,zhang2018asynchronous} and attention/self-attention (similar to our mixing function) \citep{bahdanau2014neural,vaswani2017attention} have strong similarities to our work, they lack the abstract simplicity and resulting generality (applicable to any state-dependent function) of ThinkNets and, most importantly, do not consider processing and \emph{thinking} beyond the number of timesteps used in training.

In future work, we will investigate using state-independent functions $G(x)$ within ThinkNets. We hypothesize that it is possible to transform a state-independent function such as Convolutional Neural Networks (CNN) --- commonly used in vision tasks --- via the addition of a recurrent layer near the input of these networks, making the subsequent layers state-dependent. CNNs can be over a hundred layers deep \citep{he2016deep} and ThinkNets could transform this deep processing into a shallower, iterative reasoning task.

\section{The Delta Loss}

Having defined a ThinkNet, one can simply evaluate its loss after $T$ timesteps. We propose an alternative, the Delta Loss ($\Delta \mathcal{L}$):

\ifx
\begin{align*}
  \textit{DeltaLoss}(\textit{Loss}, \textit{TN}(F,.,.,T)) =& \sum_{t=2}^T{\textit{Loss}(\textit{TN}(F,.,.,t))-\textit{Loss}(\textit{TN}(F,.,.,t-1))} \\
  & +\textit{max}_{1\leq t \leq T}(\textit{Loss}(\textit{TN}(F,.,.,t))) \\
   =& \textit{Loss}(\textit{TN}(F,.,.,T))-\textit{Loss}(\textit{TN}(F,.,.,1)) \\
  & +\textit{max}_{1\leq t \leq T}(\textit{Loss}(\textit{TN}(F,.,.,t)))
\end{align*}
\fi

\begin{equation}\label{eq:delta-loss}
\begin{aligned}
  \Delta\mathcal{L}(\mathcal{L}, \textit{TN}(F,.,.,T)) =& \left(\sum_{t=1}^T{\mathcal{L}(\textit{TN}(F,.,.,t+1))-\mathcal{L}(\textit{TN}(F,.,.,t))}\right) + \left(\underset{1\leq t \leq T}{\textit{max}}\mathcal{L}(\textit{TN}(F,.,.,t))\right) \\
   =& \mathcal{L}(\textit{TN}(F,.,.,T))-\mathcal{L}(\textit{TN}(F,.,.,1)) +\underset{1\leq t \leq T}{\textit{max}}\mathcal{L}(\textit{TN}(F,.,.,t))
\end{aligned}
\end{equation}

The first term in Equation \ref{eq:delta-loss} is the sum of all differences between subsequent losses at timesteps $t$ and $t+1$. Descending the gradient of this term encourages the model to maximize the rate with which its loss decreases at each timestep. That is, instead of encouraging the model to produce a satisfactory loss after $T$ iterations, we are encouraging it to \emph{improve its solution at each timestep}. Our hypothesis is that this should promote \emph{convergence}: models trained with Delta Loss should be able to extend their computation to more timesteps and yield better and better results. In contrast, training models only with the final $\mathcal{L}(\textit{TN}(F,.,.,T))$ could allow for poor intermediate states, which in turn could lead to divergence if they are used beyond $T$ timesteps. An interesting property of the sum-of-differences in \cref{eq:delta-loss} is that it is telescopic: all but the first and the last terms cancel out, resulting in $\mathcal{L}(TN(F,.,.,T))-\mathcal{L}(TN(F,.,.,1))$. This renders the overall equation much simpler, although no more efficient as one would still have to iterate over all losses to compute the max. loss term.

Naturally, the model is free to learn to manipulate its input in order to artificially produce a very high loss at some point, only to decrease it and collect a large delta. This is prevented by adding the maximum loss over all iterations as an additional term to \cref{eq:delta-loss}.

A trivial extension to the Delta Loss is to make the maximum loss term in \cref{eq:delta-loss} periodic: $\underset{(t-1) \bmod p = 0}{\textit{max}}\mathcal{L}(\textit{TN}(F,.,.,t))$, where $p$ is the period. This allows the state to diverge in between periodic checkpoints. At test time, the network would be evaluated using ThinkNet timesteps which are multiples of $p$. We leave the evaluation of this extension for future work.

\begin{figure}[H]
\centering
\begin{tikzpicture}

    \foreach \i in {1,...,4}
    {
        \node[rectangle, draw, minimum height=1.25cm, rounded corners=.25cm, fill=red!20!white!80, rotate=90] (unrol-\i) at (3*\i,0) {$F(x,s)$};
    }
    
    \node[rectangle, draw, minimum width=1cm, minimum height=1cm, rounded corners=.25cm, fill=red!20!white!80] (X) at (3*2.5,-2.5) {$x$};
    
    \path[->, >=stealth,  thick] (X) edge [bend left=12] node [xshift=-0cm,yshift=-0.45cm] {}  (unrol-1);
    \path[->, >=stealth,  thick] (X) edge [bend left=6] node [xshift=-0cm,yshift=-0.45cm] {}  (unrol-2);
    \path[->, >=stealth,  thick] (X) edge [bend left=-6] node [xshift=-0cm,yshift=-0.45cm] {}  (unrol-3);
    \path[->, >=stealth,  thick] (X) edge [bend left=-12] node [xshift=-0cm,yshift=-0.45cm] {}  (unrol-4);
    
    \foreach \i in {1,...,4}
    {
        \node[rectangle, draw, minimum width=.75cm, minimum height=.75cm, rounded corners=.25cm, fill=red!20!white!80] (loss-unit-\i) at (3*\i,3.5) {$\mathcal{L}$};
    }
    
    \foreach \i in {1,...,3}
    {
        \node[rectangle, draw, minimum width=.75cm, minimum height=.75cm, rounded corners=.25cm, fill=red!20!white!80] (M-unit-\i) at (3*\i+0.75,2.5) {$M$};
    }
    
    \foreach \i in {1,...,4}
    {
        \draw [ ->, >=stealth,  thick, red!50!black!50] (unrol-\i) -- (loss-unit-\i);
    }
    
    \foreach \i in {1,...,3}
    {
         \foreach \j in {\i,...,3}
         {
            \path[ ->, >=stealth, thick, red!50!black!50] (unrol-\i) edge [bend left=12] node [xshift=-0cm,yshift=-0.45cm] {}  (M-unit-\j);
         }
    }
    
     \path[ ->, >=stealth, thick, blue!50!black!50] (0.5,2) edge [bend right=15] node [xshift=.75cm,yshift=-0.25cm] {$s^{(0)}$}  ($(unrol-1) + (-0.75,0)$);
    
    \foreach \i in {1,...,3}
    {
        \pgfmathparse{\i+1}
        \xdef\iplusone{\pgfmathresult}
        
        \path[ ->, >=stealth, thick, blue!50!black!50] (M-unit-\i) edge [bend right=15] node [xshift=.75cm,yshift=-0.25cm] {$s^{(\i)}$}  ($(unrol-\iplusone) + (-0.75,-0.5)$);
    }
    
    \node[] (max-loss) at (0,4.5) {$\underset{t=1 \dots 4}{\max}\mathcal{L}^{(t)}$};
    
    \foreach \i in {1,...,4}
    {
        \node[] (loss-\i) at (3*\i,4.75) {$\mathcal{L}^{(\i)}$};
        \draw [->, >=stealth,  thick] (loss-unit-\i) -- (loss-\i);
        \path[ ->, >=stealth, thick] (loss-\i) edge [bend left=10] node [xshift=.75cm,yshift=-0.25cm] {}  (max-loss);
    }
    
    \foreach \j [remember=\j as \i (initially 1)] in {2,...,4}
    {
        \node[] (delta-\i) at (3*\i+1.5,5.5) {$\Delta(\mathcal{L}^{(\j)}, \mathcal{L}^{(\i)})$};
        \draw [->, >=stealth,  thick] (loss-\i) -- (delta-\i);
        \draw [->, >=stealth,  thick] (loss-\j) -- (delta-\i);
    }
    
    \node[rectangle, draw, minimum width=.75cm, minimum height=.75cm, rounded corners=.25cm,fill=green!20!white!80] (final-loss) at (3*2.5,7) {$\underset{t=1 \dots 4}{\max}\mathcal{L}^{(t)} + \displaystyle\sum_{i=1 \dots 3}\Delta(\mathcal{L}^{(t+1)}, \mathcal{L}^{(t)})$};
    \draw [->, >=stealth,  thick] (delta-1) -- (final-loss);
    \draw [->, >=stealth,  thick] (delta-2) -- (final-loss);
    \draw [->, >=stealth,  thick] (delta-3) -- (final-loss);
    \path[ ->, >=stealth, thick] (max-loss) edge [bend left=10] node [xshift=.75cm,yshift=-0.25cm] {}  ($(final-loss.center) - (2.75,0)$);

\end{tikzpicture}
\caption{A ThinkNet augmented with the Delta Loss. The base architecture is the same illustrated in Figure \ref{fig:thinknet}, but here we compute a loss $\mathcal{L}^{(t)}$ at each ThinkNet timestep $t$. Deltas $\Delta(\mathcal{L}^{(t+1)},\mathcal{L}^{(t)})$ between subsequent losses are computed, and then the final loss is defined as the max. loss $\max\mathcal{L}^{(t)}$ over all timesteps plus all deltas $\sum{\Delta(\mathcal{L}^{(t+1)},\mathcal{L}^{(t)})}$.}
\label{fig:deltaloss}
\end{figure}

\section{Final Remarks and Future Directions}

In this short paper we have introduced Think Again Networks. ThinkNets can be seen as an abstraction which extends any state-dependent function $F(x,s)$ by adding an additional level of recurrence to it. Inspired by the metaphor of a human problem-solver attacking a task on many ``passes'', we introduce the idea of augmenting a model with a level of recurrence which forces it to re-process the same input multiple times. In the same way that humans can learn to reason about all their previous answers when attacking a problem an additional time, we propose training a ``mixing function'' to update the state of the model with all of its previous outputs to prepare it for an additional pass. We argue that this proposal makes sense from the perspective both of humans and deep learning models, which may benefit from \emph{knowing in advance an internal representation of the input before processing it}. 

We additionally propose the companion concept of the Delta Loss to encourage a ThinkNet model to learn to \emph{improve its loss at each timestep}, which contrasts with the usual approach of training recurrent models to minimize a final loss. We hypothesize that training recurrent models with the Delta Loss could promote convergence, enabling them to extend their computation for more timesteps than they were trained on.

We wish to apply the framework to various recurrent architectures such as neural sentence encoders and graph neural networks. We also hypothesize that Think Again Networks can be applied to state-independent functions as well via a suitable transformation to state-dependent functions, for example via the addition of a recurrent layer to in feed-forward neural networks such as CNNs. 

\section*{Acknowledgments}
This research was partly supported by the Brazilian Research Council (CNPq) and Coordenação de Aperfeiçoamento de Pessoal de Nível Superior (CAPES) -- Finance Code 001.%

\medskip

\small

\bibliographystyle{unsrtnat}
\bibliography{biblio}

\end{document}